\documentclass[10pt,twocolumn,letterpaper]{article}

\usepackage{iccv}
\usepackage{times}
\usepackage{epsfig}
\usepackage{booktabs}       
\usepackage{tabularx}
\usepackage{multirow}
\usepackage[accsupp]{axessibility}
\usepackage{subcaption}
\usepackage{graphicx}
\usepackage{amsmath}
\usepackage{url}
\usepackage{amssymb}
\usepackage{bm}
\usepackage[normalem]{ulem}
\useunder{\uline}{\ul}{}

\usepackage[pagebackref=true,breaklinks=true,letterpaper=true,colorlinks,bookmarks=false]{hyperref}

\iccvfinalcopy 

\def\iccvPaperID{9515} 

\ificcvfinal\fi

\begin{document}


\title{TMA: Temporal Motion Aggregation for Event-based Optical Flow}
\author{Haotian Liu\textsuperscript{1} \qquad
Guang Chen\textsuperscript{1}\thanks{corresponding author: guangchen@tongii.edu.cn} \qquad
Sanqing Qu\textsuperscript{1} \qquad
Yanping Zhang\textsuperscript{1} \\
Zhijun Li\textsuperscript{1} \qquad
Alois Knoll\textsuperscript{2} \qquad
Changjun Jiang\textsuperscript{1} \\
{\small \textsuperscript{1}Tongji University \qquad 
\textsuperscript{2}Technical University of Munich}}

\maketitle
\ificcvfinal\thispagestyle{empty}\fi

\begin{abstract}
    Event cameras have the ability to record continuous and detailed trajectories of objects with high temporal resolution, thereby providing intuitive motion cues for optical flow estimation.
    Nevertheless, most existing learning-based approaches for event optical flow estimation directly remould the paradigm of conventional images by representing the consecutive event stream as static frames, ignoring the inherent temporal continuity of event data. 
    In this paper, we argue that temporal continuity is a vital element of event-based optical flow and propose a novel Temporal Motion Aggregation (TMA) approach to unlock its potential. 
    Technically, TMA comprises three components: an event splitting strategy to incorporate intermediate motion information underlying the temporal context, a linear lookup strategy to align temporally fine-grained motion features and a novel motion pattern aggregation module to emphasize consistent patterns for motion feature enhancement.
    By incorporating temporally fine-grained motion information, TMA can derive better flow estimates than existing methods at early stages, which not only enables TMA to obtain more accurate final predictions, but also greatly reduces the demand for a number of refinements. 
    Extensive experiments on DSEC-Flow and MVSEC datasets verify the effectiveness and superiority of our TMA. Remarkably, compared to E-RAFT, TMA achieves a 6\% improvement in accuracy and a 40\% reduction in inference time on DSEC-Flow. Code will be available at \url{https://github.com/ispc-lab/TMA}.
   
\end{abstract}
\vspace{-0.1in}
\section{Introduction}

\begin{figure}[t]
    \centering
    \vspace{-0.1in}
    \includegraphics[width=0.99\linewidth]{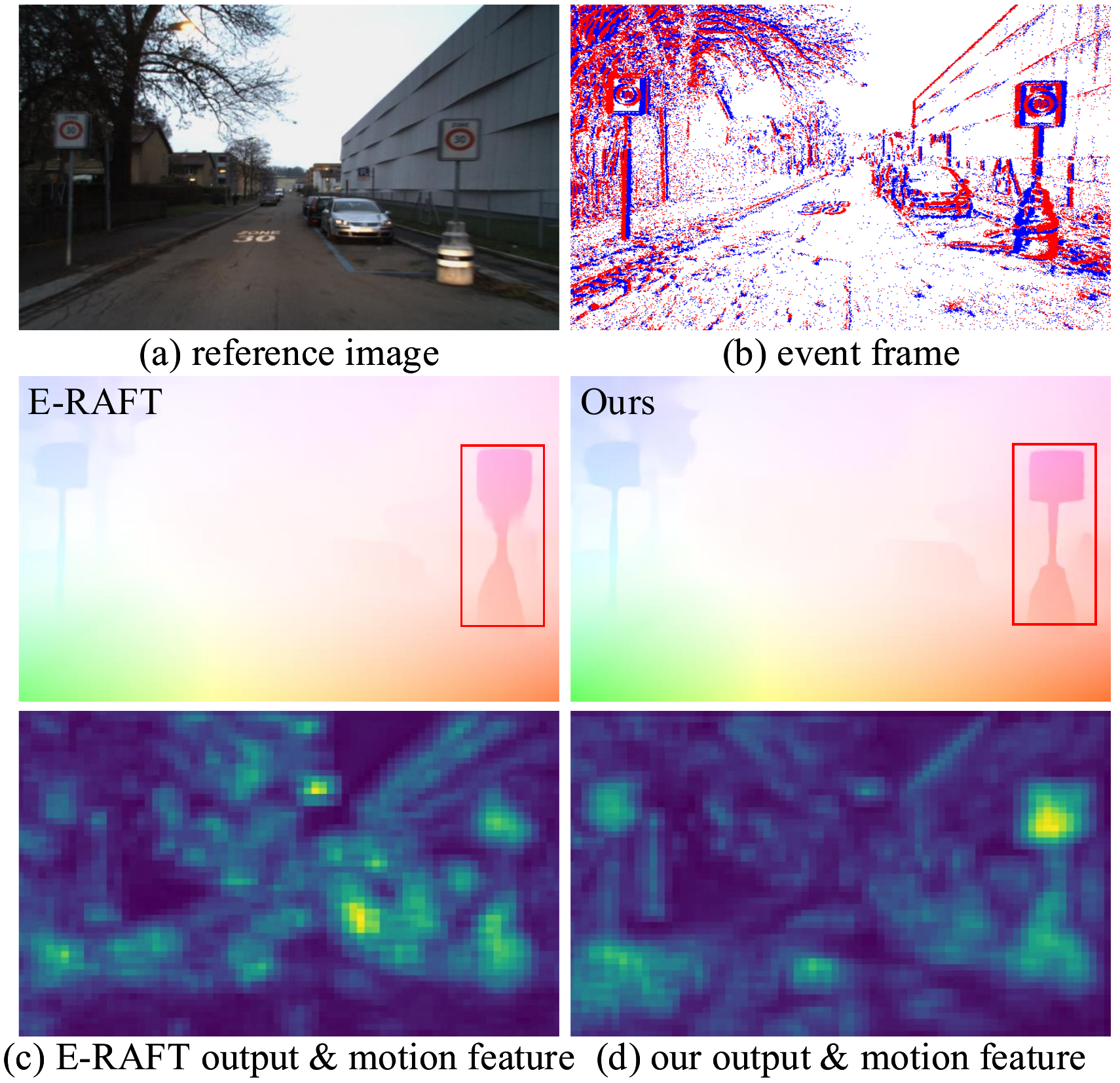}
    \vspace{-0.1in}
    \caption{\textbf{Accuracy illustrations.} (a) Reference image from DSEC-Flow. (b) Corresponding event frame. (c) The final flow prediction and visualization of the motion feature from E-RAFT~\cite{eraft}. (d) The final flow prediction and visualization of the motion feature from our proposed method. We visualize the motion feature at the initial stage by taking the average across channels. E-RAFT fails to generate informative motion features and results in blurred boundaries. In contrast, our method utilizes temporally fine-grained motion information to address the information scarcity issue in motion features, generating high-quality predictions.}
    \vspace{-0.25in}
    \label{fig:kaitou}
\end{figure}

Optical flow aims to compute the velocity of objects on the image plane without geometry prior and is a fundamental topic in event-based vision~\cite{survey, dlev}. It plays an important role in many applications, such as ego-motion estimation~\cite{yeego,zhuego}, image reconstruction~\cite{ssl}, and video frame interpolation~\cite{timelens,timelens++}. Recently, learning-based methods~\cite{pwcnet, IRR, raft} have dominated frame-based optical flow by employing correlation volumes to derive motion features for flow regression.
Inspired by this, several event-based approaches~\cite{eraft, TIP} adopt a similar paradigm by converting the consecutive event stream into grid-like representations. 
Despite encouraging progress, these methods still suffer from non-negligible limitations, due to the great differences between event data and conventional images. Distinct from dense and colorful conventional (RGB) images, event data features spatial sparsity and a lack of intensity information. It is prone to result in close matching scores and invalid regions in the correlation volume. Consequently, less informative motion features are derived and regress inaccurate predictions. Figure~\ref{fig:kaitou} illustrates this issue in E-RAFT~\cite{eraft}.

As the saying goes, each coin has two sides. Though event cameras encode visual information in a sparse manner, they are capable of capturing continuous and detailed object trajectories with high temporal resolution, thereby providing rich motion cues for optical flow estimation. 
Several model-based methods~\cite{CM, secrets} benefit from the temporal continuity by warping events along point trajectories, achieving decent performance. 
Drawing on this observation, we contend that the temporal continuity is a vital element of event-based optical flow and propose to unleash its potential within a learning-based framework.

To materialize our idea, we propose a novel Temporal Motion Aggregation (TMA) approach to explore the inherent temporal continuity of event data. Technically, we revamp the dominant learning-based architectures with three distinct components.
We first introduce an event splitting strategy for temporally-dense correlation volumes computation. By splitting the event stream into multiple segments and extracting their features, feature similarities are compared between the first feature and all others, which record rich intermediate motion information.
Aware of that correlation volumes record motions between different time spans, we then design a linear lookup strategy to sample each correlation volume based on the corresponding flow estimate to encode motion features. As a result, fine-grained motions are warped to the same pixels across motion features.  
Furthermore, we consider the incorrect motion patterns in intermediate motion features due to the manual lookup. Therefore, we propose a novel pattern aggregation module for motion feature enhancement by aggregating consistent patterns between each intermediate motion feature and the last one (without manual lookup).
Thanks to the incorporation of temporally fine-grained motion information, TMA can effectively address the information scarcity in motion features and thus generate good flow estimates at early stages, which not only enable TMA to obtain accurate flow predictions, but also greatly reduces the demand for a number of time-consuming refinements. 

We evaluate our TMA on DSEC-Flow~\cite{dsec} and MVSEC~\cite{mvsec} datasets. Extensive experiments demonstrate the superior advantages of TMA.

In summary, our main contributions are as follows:
\begin{itemize}
\vspace{-0.05in}
    \item We argue that the temporal continuity of event data is a vital element of event-based optical flow and propose to unlock its potential in a learning-based framework.
    \vspace{-0.05in}
    \item We propose a novel Temporal Motion Aggregation (TMA) approach, which comprises three components: an event splitting strategy, a linear lookup strategy and a motion pattern aggregation module.
    \vspace{-0.05in}
    \item The incorporation of temporally fine-grained motion information enables TMA to achieve high accuracy while maintaining high efficiency. Compared with E-RAFT, TMA achieves a 6\% improvement in accuracy and a 40\% reduction in inference time on DSEC-Flow.

\end{itemize}

\section{Related Work}

\begin{figure*}[ht]
    \centering
    \includegraphics[width=0.99\linewidth]{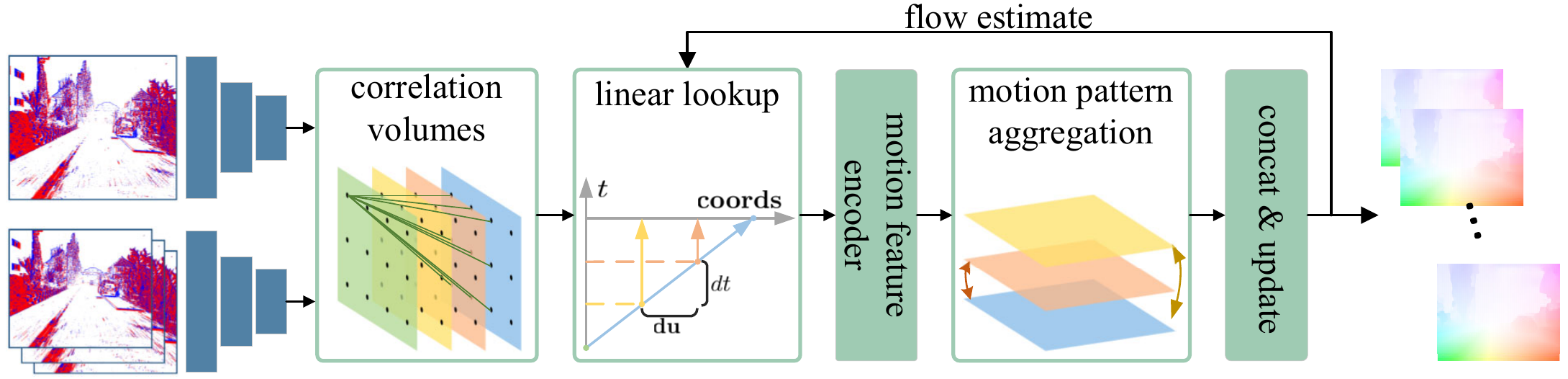}
    \vspace{-0.1in}
    \caption{\textbf{The overall architecture of TMA}. First, we split the event stream into segments and extract features with a share-weights feature extractor. Then, regarding the first feature as reference, we correlate all other features to compute temporally-dense correlation volumes, which contain rich embedded motion information. Next, aware of the different time spans of correlation volumes, we look up each correlation volume based on a corresponding flow estimate. All sampled correlation maps are delivered into a share-weights motion feature encoder to derive motion features. We further propose a novel motion pattern aggregation module to enrich the spatial information of motion features. By concatenating the enhanced motion features and utilizing several refinements, high-quality flow predictions are generated.}
    \vspace{-0.2in}
    \label{fig:overall}
\end{figure*}

\textbf{Learning methods for optical flow estimation}. Recently, correlation volume based approaches dominate the learning-based optical flow estimation~\cite{spynet, liteflownet2,pwcnet,IRR,VCN, improveflowpyramid}. As a milestone, PWCNet~\cite{pwcnet} summarizes a simple principle in architecture construction: pyramidal processing, warping and correlation volumes computation. Following the well-established pipeline, LiteFlowNet series~\cite{liteflownet,liteflownet2,liteflownet3} propose cascaded feature pyramids with a lightweight convolutional burden.
VCN~\cite{VCN} proposes 4D volumetric U-Net architectures to enlarge the receptive fields for correlation volumes. IRR~\cite{IRR} notes the reusability of pyramidal refinement for flow update and proposes a shared-weights refiner across all scales, which can be combined with mainstream learning methods.

However, the correlation volumes used in previous works are computed locally, which cannot fully deal with long-term motions. RAFT~\cite{raft} firstly introduces the full correlation volume to capture global visual similarities.
Separable Flow~\cite{separable} proposes a separable correlation volume module as a drop-in replacement to correlation volumes and uses non-local aggregation layers~\cite{ganet, domain} to refine the correlation volumes efficiently. Besides, transformer components show promising performance on optical flow~\cite{gmflow,gma,1Dattention}. GMA~\cite{gma} notices the underlying motion clues in the context and introduces an attention layer to aggregate hidden motions, improving the accuracy based on RAFT architecture. GMFlow~\cite{gmflow} reformulates optical flow as a global matching problem by attention blocks, suggesting a new paradigm for flow estimation.

We draw inspiration from GMA~\cite{gma} to design our motion pattern aggregation module. In contrast to GMA which aggregates hidden motion patterns in the spatial context for motion feature enhancement, we extend the pattern aggregation to the temporal dimension and leverage feature cross-similarities to enhance the motion feature.

 \textbf{Event-based optical flow estimation.} A number of traditional works remould frame-based approaches into event-based optical flow estimation. Benosman et al.~\cite{asynchronous} extends the classic Lucas-Kanade method to event-based optical flow. Liu et al.~\cite{block, EDFlow} implement event-based corner keypoint detection and novel block-matching optical flow. Gallego et al.~\cite{CM} propose a unifying contrast maximization (CM) framework and MultiCM~\cite{secrets} extends it to complex scenes. Besides, plane fitting~\cite{aperture-robust,evisualflow}, filter banks~\cite{eofdection} and time surface matching~\cite{distancePAMI} are proposed to boost performance.

More recently, learning-based methods have dominated frame-based optical flow, inspiring attempts at event-based optical flow. Early learning methods~\cite{evflownet,zhuego,yeego} represent the event stream as a static frame and utilize the architecture of U-Net~\cite{unet} to predict optical flow, which can only address small motions. Mainstream works~\cite{STE,TIP,eraft} introduce two consecutive event frames and follow the correlation volume based pipeline (PWCNet and RAFT), achieving higher accuracy on large-scales datasets~\cite{mvsec, dsec}. E-RAFT~\cite{eraft, continuous} extends RAFT~\cite{raft} and introduces full correlation volume into event-based vision. DCEIFlow~\cite{TIP} combines conventional images and event data to boost higher accuracy. Note that our work and STE-FlowNet~\cite{STE} share a similar motivation to introduce temporal continuity with fine-grained discretization. However, STE-FlowNet takes a frame-by-frame manner~\cite{IRR} to update flow, leading to an insufficient inference process.

Contrary to previous learning-based methods, we do not merely adopt the classic correlation volume based pipeline. Instead, we regard the temporal continuity of event data as a vital component of optical flow and incorporate the intermediate motion information into a learning-based framework, which boosts higher accuracy and efficiency. 

\section{Methodology}
Given an input event stream $\{\mathcal{E}\}$ from $t_0$ to $t_1$, the goal of optical flow is to estimate a dense displacement field $\bm{\mathrm{u}}: \mathbb{R}^2 \rightarrow \mathbb{R}^2$ which maps each per-pixel $(x, y)$ at $t_0$ to its corresponding coordinate $(x^\prime, y^\prime)$ at $t_1$. To address this, mainstream learning-based approaches compute correlation volumes from two event frames as a matching prior. Such principle inspires several methods~\cite{eraft,TIP} by squeezing event data into consecutive frames. Unfortunately, the characteristics of event data, spatial sparsity and lack of intensity records, are prone to result in close matching scores and invalid regions in the correlation volume. Consequently, less informative motion features are derived and regress inaccurate predictions. Here we address the information scarcity of motion features from the attractive temporal continuity of event data.
Temporally fine-grained motion information in event data is sufficient to supplement the spatial richness of motion features, yielding high-quality flow predictions.

In the following, we first describe the preliminary and event representation, and then present the detailed architecture of the proposed TMA.

\subsection{Preliminary and Event Representation}
\label{sec:pre}
Event cameras record light intensity changes in an asynchronous way. Each pixel serves as an independent trigger and generates an event instantly whenever the log intensity changes exceeding a threshold $C$. An event $\bm{e}_k = (\bm{x}_k, p_k, t_k)$ is a triplet containing triggered pixel location $\bm{x}_k = (x_k,y_k)^T$, timestamp $t_k$ and signed intensity change $(p_k=\pm 1)$.
\par To ease the feature extraction and preserve the temporal continuity of the event data, we firstly transform the event stream into a 3D volume $\bm{V}(x, y, t)$ along the temporal dimension following existing works~\cite{zhuego, eraft, ssl}. Given an event stream $\{\mathcal{E}\}$, $\bm{V}(x, y, t)$ is generated as:
\begin{align}
&t_{i}^{*} =(B-1)(t_{i}-t_{1}) /(t_{N_e}-t_{1}) \\
&\bm{V}(x, y, t) =\sum_{i} p_{i} k_{b}(x-x_{i}) k_{b}(y-y_{i}) k_{b}(t-t_{i}^{*}) \\
&k_{b}(a) =\max (0,1-|a|),
\end{align}
where $t_{i}^{*}$ represents the $i$-th time bin. $B$ and $N_e$ represent the length of time bins and the number of events, respectively. $k_{b}(a)$ is the bilinear interpolation function.

\subsection{TMA Architecture}
We develop our framework based on the successful RAFT architecture~\cite{raft}. Figure~\ref{fig:overall} summarizes the architecture of the proposed TMA, which consists of three components, \textit{an event splitting strategy}, \textit{a linear lookup strategy} and \textit{a motion pattern aggregation module}. With the proposed method, we successfully unlock the potential of temporal continuity to predict event-based optical flow within a learning-based framework.

\textbf{Event splitting.} Instead of viewing the continuous event stream as a static frame which tends to be a matter of common knowledge in recent works~\cite{zhuego, eraft, TIP}, we prefer to identify it as a high frame rate video clip, which records continuous and detailed trajectories of objects, namely, provides intuitive motion cues for optical flow estimation. To introduce rich intermediate motion information into our method, we first propose an event splitting strategy, which splits the event stream into multiple segments. Therefore, more intermediate event frames are retained for correlation volume computation. 

In particular, given an event stream $\{\mathcal{E}\}$ from $t_0$ to $t_1$, we split it into $g$ segments equally and the time span between adjacent segments is $dt$. Besides, we introduce an auxiliary event segment from $t_0-dt$ to $t_0$ as reference. Then, $g+1$ event representations are generated following Sec.~\ref{sec:pre}. All $g+1$ event representations are then delivered into a share-weights encoder to achieve features $\bm{F}_i\in \mathbb{R} ^{H\times W\times C}, i=0, 1,2,...,g$, where $H$, $W$ and $D$ represent the height, width and channel of features. We use all features to compute temporally-dense correlation volumes $\bm{C}_i, i=1,2,...,g$, which are given by:
\begin{equation}
    \bm{C}_i = \frac{\bm{F}_0\bm{F}_i^T}{\sqrt{D}}\in \mathbb{R} ^{H\times W\times H\times W}, i=1,2,...,g,
\end{equation}
where $\bm{F}_0$ is the first feature and $\bm{F}_i$ is the $i$-th feature derived from the $i$-th event frame. 

\begin{figure}[t]
    \centering
    \includegraphics[width=0.90\linewidth]{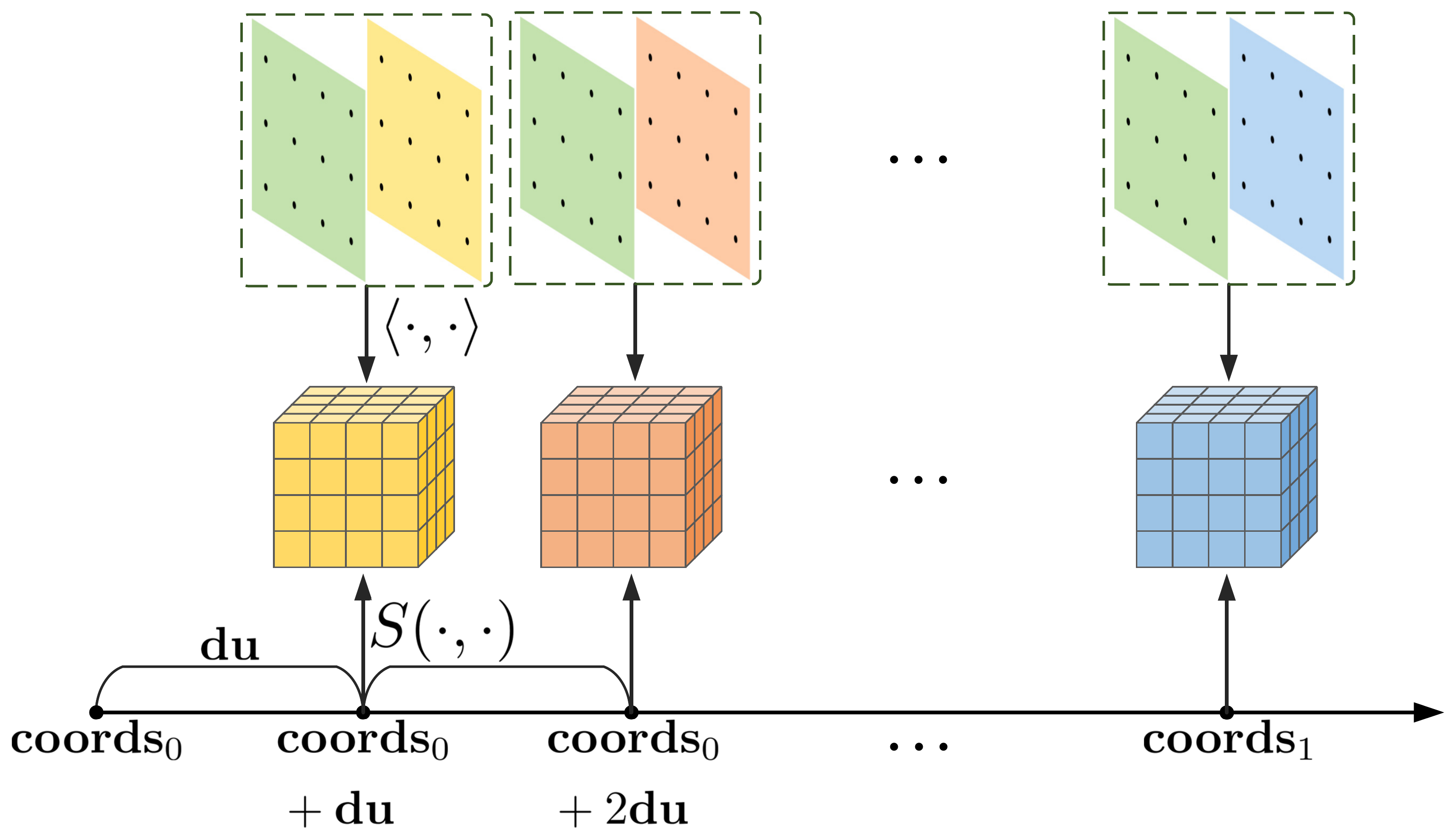}
    \vspace{-0.1in}
    \caption{\textbf{Linear lookup strategy.} We compute correlation volumes by comparing the similarity of first feature with all others, which are misaligned due to different time spans. To rectify this, we narrow down the lookup coordinates linearly to warp correlation maps to spatially aligned pixels.}
    \vspace{-0.2in}
    \label{fig:linear}
\end{figure}

Temporally-dense correlation volumes compare feature similarities across multiple time spans, enabling refined records of relative motions.
This construction is vastly different from the classic full correlation volume that only contains feature similarities between a fixed time span.

\textbf{Linear lookup.} Recall that each correlation volume $\bm{C}_i$ is achieved by correlating $\bm{F}_0$ and $\bm{F}_i$, which means the recorded relation motions are misaligned in time.
While our goal is to predict optical flow from $t_0$ to $t_1$, we hope that all sampled correlation maps could present the relative motions within the same time span. 
Therefore, based on that correlation volumes are evenly constructed in time and magnitudes of optical flow change linearly over a short period, we narrow down lookup coordinates in a linear fashion to match the timestamp of each correlation volume.

Figure~\ref{fig:linear} displays the linear lookup strategy. In detail, given current flow estimate $\bm{\mathrm{u}}$, we split it into $g$ segments to sample each correlation volume. The linear lookup operation can be expressed as:
\begin{align}
    &\bm{\mathrm{du}} = (\bm{\mathrm{coords}}_1 - \bm{\mathrm{coords}}_0)/g = \bm{\mathrm{u}}/g\\
    &\bm{\mathrm{corr}}_i = S(\bm{C}_i, \bm{\mathrm{coords}}_0 + i\bm{\mathrm{du}}), i=1,2,...,g,
\end{align}
where $\bm{\mathrm{coords}}_0$ is the initialized coordinates at $t_0$ (reference time), $S(\cdot, \cdot)$ represents the lookup operation, $\bm{\mathrm{coords}}_i$ are lookup coordinates for $\bm{C}_i$ and $\bm{\mathrm{corr}}_i$ is the correlation map sampled from $\bm{C}_i$. After delivering all correlation maps into a share-weights motion feature encoder, aligned motion features are achieved.

From the perspective of image contrast~\cite{CM}, the standard lookup is responsible for warping possible matching points between two images to the same pixel. Our linear lookup extends it by warping matching points of a temporally fine-grained motion to the same pixel. As a result, similar motions are presented at same pixels across correlation maps, showing high contrast. 

\begin{figure}[t]
    \centering
    \includegraphics[width=0.88\linewidth]{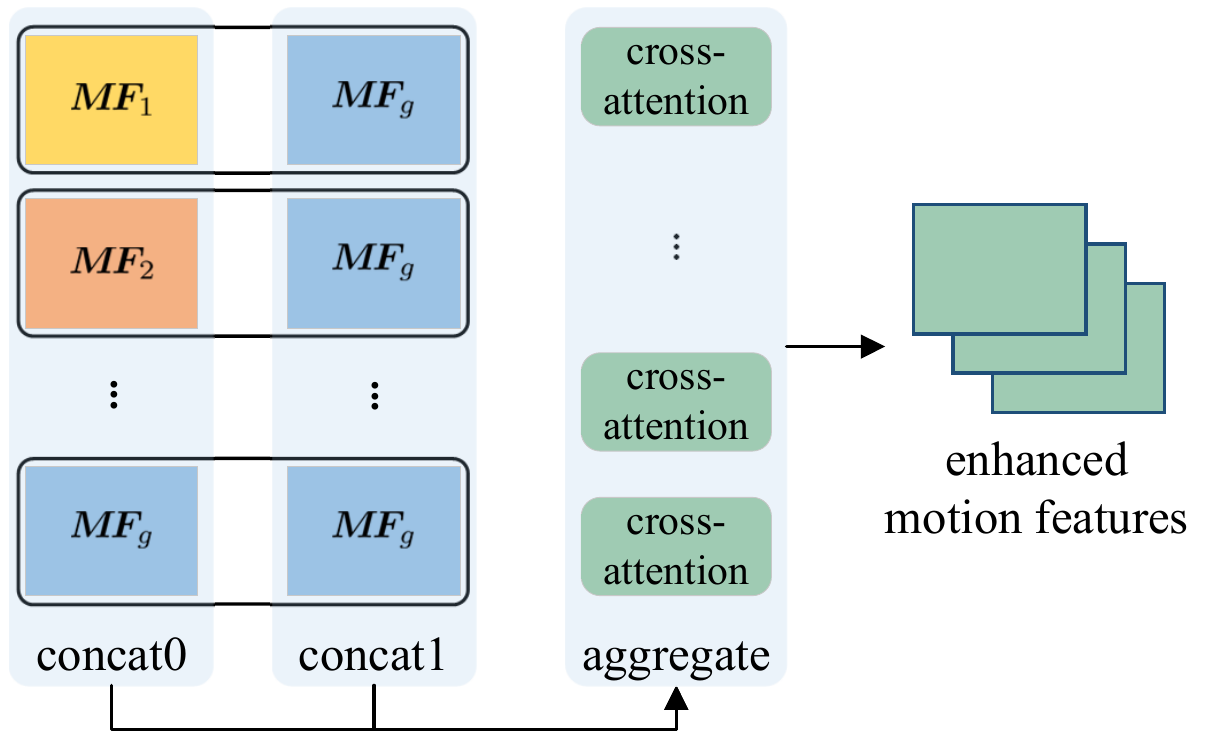}
    \vspace{-0.1in}
    \caption{\textbf{Motion pattern aggregation.} We pair each intermediate motion feature to the last motion feature. Then each pair is sent to cross-attention blocks to aggregate consistent patterns. Hence, each intermediate motion feature carries information from the last one and consistent patterns receive emphasis.}
    \vspace{-0.2in}
    \label{fig:MPA}
\end{figure}

\begin{figure*}[t]
    \centering
    \includegraphics[width=0.995\linewidth]{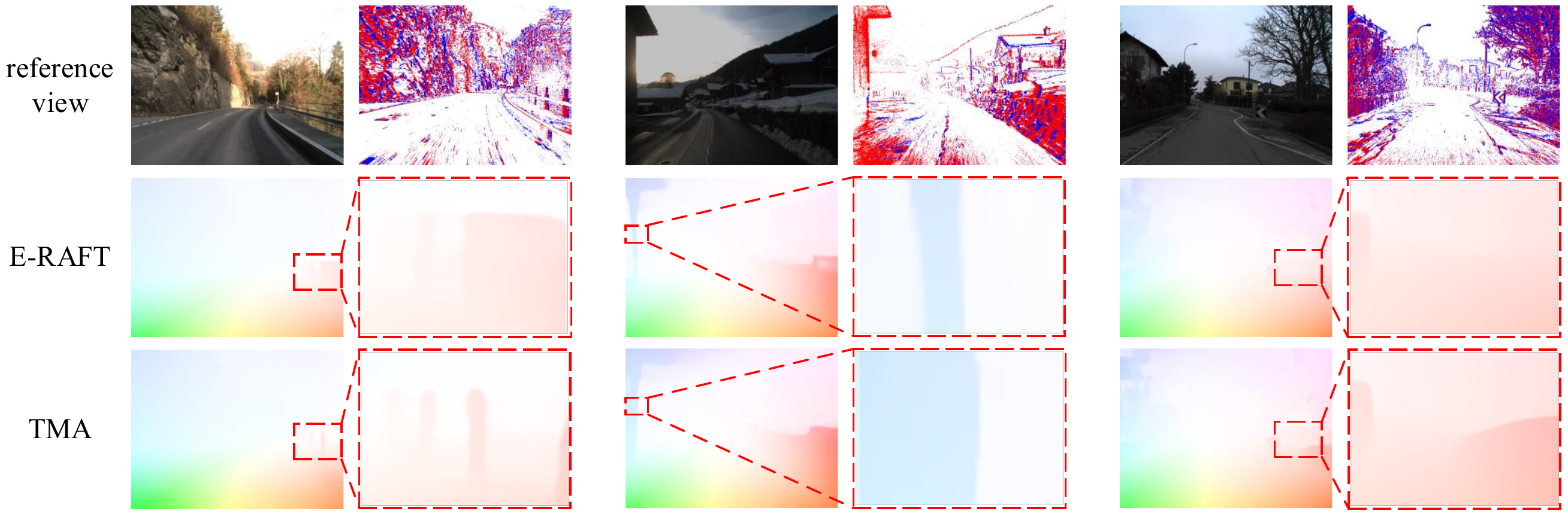}
    \vspace{-0.1in}
    \caption{\textbf{Qualitative examples of DSEC-Flow.} For each two columns, we show the reference image and event frame and compare our method with a state-of-the-art baseline E-RAFT~\cite{eraft}. Significant improvements are highlighted by red boxes.}
    \vspace{-0.15in}
    \label{fig:qua_dsec}
\end{figure*}

\textbf{Motion pattern aggregation.} A standard operation for next step is to concatenate the motion features as a whole to update flow and the improvements made so far already alleviate the information scarcity in motion features. 
However, we notice that the concatenation operation treats all motion features equally, which means the majority of motion features, intermediate ones, dominate the flow regression. 
Due to the linear optical flow assumption and manual lookup operation, these motion features are not perfectly correct.
A natural thought is to put more emphasis on the last motion feature that is without manual lookup and retain the patterns in intermediate motion features that are consistent with the last motion feature. 
To this end, we balance the weights between the last motion feature and others through cross-similarities computed by transformers~\cite{attention,gma}.

Figure~\ref{fig:MPA} shows the overview of motion pattern aggregation module. Given a set of motion features $\bm{M\!F}_i \in \mathbb{R} ^{H\times W\times D'}, i=1,2,...,g$, we pair each intermediate motion feature to the last one. Then each pair $(\bm{M\!F}_i, \bm{M\!F}_g)$ is sent to cross-attention blocks to compare cross-similarities and aggregate consistent patterns. The enhanced motion feature $\widehat{\bm{M\!F}}_i$ is obtained as: 
\begin{align}
    &\bm{A} = \operatorname{Attention}(\bm{M\!F}_i W^Q, \bm{M\!F}_g W^K, \bm{M\!F}_g W^V)\\
    &\widehat{\bm{M\!F}}_i = \bm{M\!F}_i + \operatorname{MLP}([\bm{M\!F}_i, \bm{A} W^O]),
\end{align}
where $\operatorname{Attention}(Q, K, V)=\operatorname{softmax}\left(\frac{Q K^T}{\sqrt{d_k}}\right) V$, $d_k$ is the dimension of $K$. $W^Q$, $W^K$, $W^V$ are projection matrices. $\operatorname{MLP}(\cdot)$ is multi-layer perceptron and $W^O$ is another projection matrix. We omit the normalization functions in the manuscript for simplicity. Motion pattern aggregation module does not change the dimension of motion features, so multiple cross-attention layers can be stacked easily.

Once motion features are enhanced by the motion pattern aggregation module, each $\widehat{\bm{M\!F}}_i$ carries motion patterns that are consistent with $\bm{M\!F}_g$. Namely, the last motion feature receives high attention and consistent patterns in motion features are fully utilized.

Above three components are seamlessly integrated into a unified pipeline: incorporating intermediate motion information, warping matching motions and emphasizing consistent motion patterns. 
Attributed to the innovative application of temporal continuity to a learning-based framework, our proposed method is capable of overcoming the information scarcity issue in motion features and generating high-quality flow estimates even at an early refinement stage. This enables our method to achieve accurate final predictions, while reducing the demand for a number of refinements.

\subsection{Supervision}
Following RAFT~\cite{raft}, we choose $L_1$ distance between the predictions and ground-truth as the supervision signal for our model. The loss function $\mathcal{L}$ is formally defined as:
\begin{equation}
    \mathcal{L}=\sum_{j=1}^{N} \gamma^{N-j}\left\|\bm{\mathrm{u}}^{gt}-\bm{\mathrm{u}}^{j}\right\|_{1},
\end{equation}
where $\bm{\mathrm{u}}^{gt}$ denotes the flow ground-truth and $\bm{\mathrm{u}}^{j}$ the flow prediction at $j$-th stage. $N$ is the total refinement stages. The weight $\gamma$ balances different stages of predictions.

\section{Experiments}
\textbf{Datasets and evaluation setup.} 
Following previous works~\cite{eraft, zhuego,STE}, we provide experimental details and extensive comparison results on two event-based datasets DSEC-Flow~\cite{dsec} and MVSEC~\cite{mvsec}. For DSEC-Flow, we train models on the official training set and evaluate DSEC-Flow test set on the public benchmark. 
For MVSEC, we conduct two kinds of experiments. Basically, following previous methods~\cite{evflownet, secrets}, we train methods on outdoor\_day2 sequence with ground-truth corresponding to a time interval of $dt=1$ and $dt=4$ gray images and evaluate 800 frames of outdoor\_day1 and three indoor\_flying sequences. Equally importantly, considering the supervised learning setting of our method and the large gap between outdoor\_day and indoor\_flying data, we extend the range of training data by including one indoor\_flying sequence each time to further boost the accuracy of supervised learning methods.

\textbf{Metrics.} The end-point-error (EPE) is used as core metric for prediction accuracy for both DSEC-Flow and MVSEC. To evaluate the robustness against large displacements, we compute the percentage of pixels with EPE greater than $M$ pixels ($M$PE, $M=1, 3$) for DSEC-Flow. Percentage of pixels with EPE above 3 and 5\% of the magnitude of the flow ground-truth (\% Outlier) are presented for MVSEC. On DSEC-Flow, metrics are measured over pixels with valid ground-truth. While on MVSEC, metrics are measured over pixels with valid ground-truth and at least one triggered event.

\textbf{Implementation details.} The proposed TMA is implemented using PyTorch. We set the event segment $g=5$. For experiments on DSEC-Flow, the channel of event representation $B$ for each segment is set to 3. For MVSEC, $B$ is set to 1 when $dt=1$ and 3 when $dt=4$, respectively. The sum of channels keeps the same with the state-of-the-art E-RAFT~\cite{eraft}. We only stack 1 layer of motion pattern aggregation module. Other components are basically identical to RAFT’s model, including feature extractor, motion feature encoder and flow updater. Note that we use a small feature extractor whose final channel is 128. We predict 6 levels of optical flow and the weight $\gamma$ is set to 0.8 for supervision. In training, we exploit AdamW~\cite{adamw} optimizer and One-cycle learning rate scheduler~\cite{onecycle}. We train models on DSEC-Flow for 200K steps and MVSEC for 100K steps, both with a learning rate of 0.0002 and a batch size of 6.

\begin{table}[t]
\centering
\addtolength{\tabcolsep}{1.0pt}
\resizebox{0.47\textwidth}{!}{
    \begin{tabular}{llcccc}
    \toprule
    &Methods  & EPE & 1PE & 3PE& Time(ms) \\ 
    \midrule
    MB & MultiCM~\cite{secrets}      & 3.47           & 76.6	        & 30.9  &-    \\
    \midrule
    \multirow{3}{*}{SL}&EV-FlowNet$^\dag$~\cite{evflownet}   & 2.32            &55.4        &  18.6 & 5   \\
    &E-RAFT~\cite{eraft}        & 0.79           & 12.5           & 2.7  & 52     \\
    &\textbf{Ours}              & \textbf{0.74}   & 	\textbf{10.9}  &\textbf{2.3}  & 30\\
    \bottomrule
    \end{tabular}}
\vspace{-0.05in}
\caption{\textbf{Results on DSEC-Flow.} Best accuracy in bold. $^\dag$ denotes the result is taken from E-RAFT~\cite{eraft}. Model-based (MB) methods need no training data; and supervised learning (SL) methods need ground-truth.}
\vspace{-0.20in}
\label{table:dsec}
\end{table}

\begin{table*}[t]
\centering
  \addtolength{\tabcolsep}{2.0pt}
  \resizebox{0.99\textwidth}{!}{
    \begin{tabular}{ll@{\extracolsep{\fill}}ccccccccc}
    \toprule
    \multicolumn{2}{l}{\multirow{2}{*}{}}   &\multirow{2}{*}{Input}& \multicolumn{2}{c}{indoor\_flying1} & \multicolumn{2}{c}{indoor\_flying2} & \multicolumn{2}{c}{indoor\_flying3} & \multicolumn{2}{c}{outdoor\_day1} \\ 
        \cmidrule(r){4-5}\cmidrule(lr){6-7}\cmidrule(lr){8-9}\cmidrule(lr){10-11}
        & & &EPE & \% Outlier   & EPE & \% Outlier   & EPE & \% Outlier   & EPE & \% Outlier   \\
    \midrule
    $dt = 1$&&&&&&&&&&\\
    
    \midrule
    \multirow{2}{*}{MB}&Brebion et al.~\cite{lowhigh}&E     &0.52 &0.10   &0.98 &5.50   &0.71 &2.10   &0.53 &0.20   \\
    &MultiCM (Burgers')~\cite{secrets}&E &\textbf{0.42} &0.10   &\textbf{0.60} &\textbf{0.59}   &\textbf{0.50} &\textbf{0.28}   &0.30 &0.10   \\
    \midrule
    \multirow{3}{*}{SSL}&EV-FlowNet~\cite{evflownet}&E       &1.03 &2.20   &1.72 &15.10  &1.53 &11.90  &0.49 &0.20   \\
    &Spike-FlowNet~\cite{spikeflownet}&E &0.84 &–      &1.28 &–      &1.11 &–      &0.49 &–     \\
    &STE-FlowNet~\cite{STE}&E            &0.57 &0.10   &0.79 &1.60   &0.72 &1.30   &0.42 &\textbf{0.00}   \\
    \midrule
    \multirow{3}{*}{USL}&Hagenaars et al.~\cite{self}&E    &0.60 &0.51   &1.17 &8.06   &0.93 &5.64   &0.47 &0.25   \\
    &Zhu et al.~\cite{zhuego}&E          &0.58 &\textbf{0.00}   &1.02 &4.00   &0.87 &3.00   &0.32 &\textbf{0.00}   \\
    &ECN~\cite{yeego}&E &-&-&-&-&-&-&0.30&0.02\\
    \midrule
    \multirow{3}{*}{SL}&E-RAFT~\cite{eraft}&E               &1.10    &5.72      &1.94    &30.79      &1.66    &25.20     &0.24 &\textbf{0.00}   \\
    &DCEIFlow~\cite{TIP}&E + I  & 0.75    &1.55       &  0.90   &  2.10     & 0.80    & 1.77      &  \textbf{0.22}   &    \textbf{0.00}   \\
    &\textbf{Ours}&E                      &1.06    & 3.63      &1.81     & 27.29      &  1.58   &23.26       &  0.25   &  0.07     \\

    \midrule
    $dt= 4$&&&&&&&&&&\\
    \midrule
    \multirow{1}{*}{MB}&MultiCM (Burgers')~\cite{secrets}&E &\textbf{1.69} &\textbf{12.95}  &\textbf{2.49} &26.35  &\textbf{2.06} &\textbf{19.03}  &1.25 &9.21   \\
    \midrule
    \multirow{3}{*}{SSL}&EV-FlowNet~\cite{evflownet}&E       &2.25 &24.70  &4.05 &45.30  &3.45 &39.70  &1.23 &7.30   \\
    &Spike-FlowNet~\cite{spikeflownet}&E &2.24 &–      &3.83 &–      &3.18 &–      &1.09 &–      \\
    &STE-FlowNet~\cite{STE}&E            &1.77 &14.70  &2.52 &\textbf{26.10}  &2.23 &22.10  &0.99 &3.90   \\
    \midrule
    \multirow{2}{*}{USL}&Zhu et al.~\cite{zhuego}&E          &2.18 &24.20  &3.85 &46.80  &3.18 &47.80  &1.30 &9.70   \\
    &Hagenaars et al.~\cite{self}&E    &2.16 &21.51  &3.90 &40.72  &3.00 &29.60  &1.69 &12.50  \\
    \midrule
    \multirow{3}{*}{SL}&DCEIFlow~\cite{TIP}&E + I             &2.08  &21.47   &3.48 &42.05  &2.51 &29.73 &0.89&3.19       \\
    &E-RAFT~\cite{eraft}&E &2.81&	40.25&	5.09&	64.19&	4.46&	57.11&0.72& 	1.12\\
    &\textbf{Ours}&E            &2.43&	29.91&	4.32&	52.74&	3.60&	42.02& \textbf{0.70}&	\textbf{1.08}   \\
    \bottomrule
    \end{tabular}
\vspace{-0.25in}
}
\caption{\textbf{Evaluation results on MVSEC~\cite{mvsec} training with outdoor\_day2 sequence.} Model-based (MB) methods need no training data; semi-supervised learning (SSL) methods use grayscale images for supervision; unsupervised learning (USL) methods only require events; and supervised learning (SL) methods need ground-truth. E means the method only uses events as input; and I means the method requires images as input.}
\label{tab:mvsec}
\end{table*}

\begin{table*}[ht]
\centering
  \addtolength{\tabcolsep}{4.0pt}
  \resizebox{0.99\textwidth}{!}{
    \begin{tabular}{llcccccccc}
    \toprule
    \multirow{2}{*}{$dt=4$}   &\multirow{2}{*}{Train Set +}& \multicolumn{2}{c}{indoor\_flying1} &\multicolumn{2}{c}{indoor\_flying2} & \multicolumn{2}{c}{indoor\_flying3} & \multicolumn{2}{c}{outdoor\_day1} \\ 
        \cmidrule(r){3-4}\cmidrule(lr){5-6}\cmidrule(lr){7-8}\cmidrule(lr){9-10}
    & & EPE & \% Outlier   & EPE & \% Outlier   & EPE & \% Outlier   & EPE & \% Outlier\\
    \midrule
    E-RAFT~\cite{eraft}  & \multirow{2}{*}{indoor\_flying1}&-   & - &2.35&26.24&1.84&18.07&0.72&1.38\\
    Ours  &      &    -        &- &\textbf{2.15} &\textbf{20.41} &\textbf{1.68}      &\textbf{14.30} & \textbf{0.64}& \textbf{0.98}   \\
    \midrule
    E-RAFT~\cite{eraft}  & \multirow{2}{*}{indoor\_flying2}&1.41 &8.14 &-&-&\textbf{1.50}&9.37&0.71&1.13   \\
    Ours  &                     &\textbf{1.32}           &\textbf{5.94}&-&-&\textbf{1.50}&\textbf{8.88}&\textbf{0.64}&\textbf{0.96}\\
    \midrule
    E-RAFT~\cite{eraft}  & \multirow{2}{*}{indoor\_flying3}&1.39 &8.38&1.77&17.52&-&-&0.72&1.46    \\
    Ours  &         &\textbf{1.33} &\textbf{5.82}&\textbf{1.67}&\textbf{12.80}&-&-&\textbf{0.67}&\textbf{1.10}\\
    \bottomrule
    \end{tabular}
    }
\vspace{-0.1in}
\caption{\textbf{Evaluation results on MVSEC~\cite{mvsec} training with outdoor\_day2 and one indoor\_flying sequence.} + means the indoor\_flying sequence to be included into training set.}
\vspace{-0.2in}
\label{tab:MVSEC_gen}
\end{table*}

\subsection{DSEC-Flow}
\textbf{Accuracy comparison.} Evaluation results on DSEC-Flow benchmark\footnote{The ground-truth of DSEC-Flow dataset is unavailable, so we use last three checkpoints to evaluate the online benchmark. \url{https://dsec.ifi.uzh.ch/uzh/dsec-flow-optical-flow-benchmark/}.} are provided in Table~\ref{table:dsec}. Compared with E-RAFT, our method improves the EPE from 0.79 to 0.74. Our method is also leading in 1PE and 3PE with a margin of 1.6 and 0.4, showing better robustness.

Qualitative results are exhibited in Figure~\ref{fig:qua_dsec}. In challenging regions such as textureless areas (the wall in second two columns, the shrubs in last two columns), existing correlation volume contains similar matching scores and fails to generate informative motion features, leading to low-quality predictions. On the contrary, our method aggregates temporally consistent motion patterns to enhance motion features, distinguishing different moving objects and generating high-quality predictions.

\begin{figure}[t]
    \centering
    \includegraphics[width=0.97\linewidth]{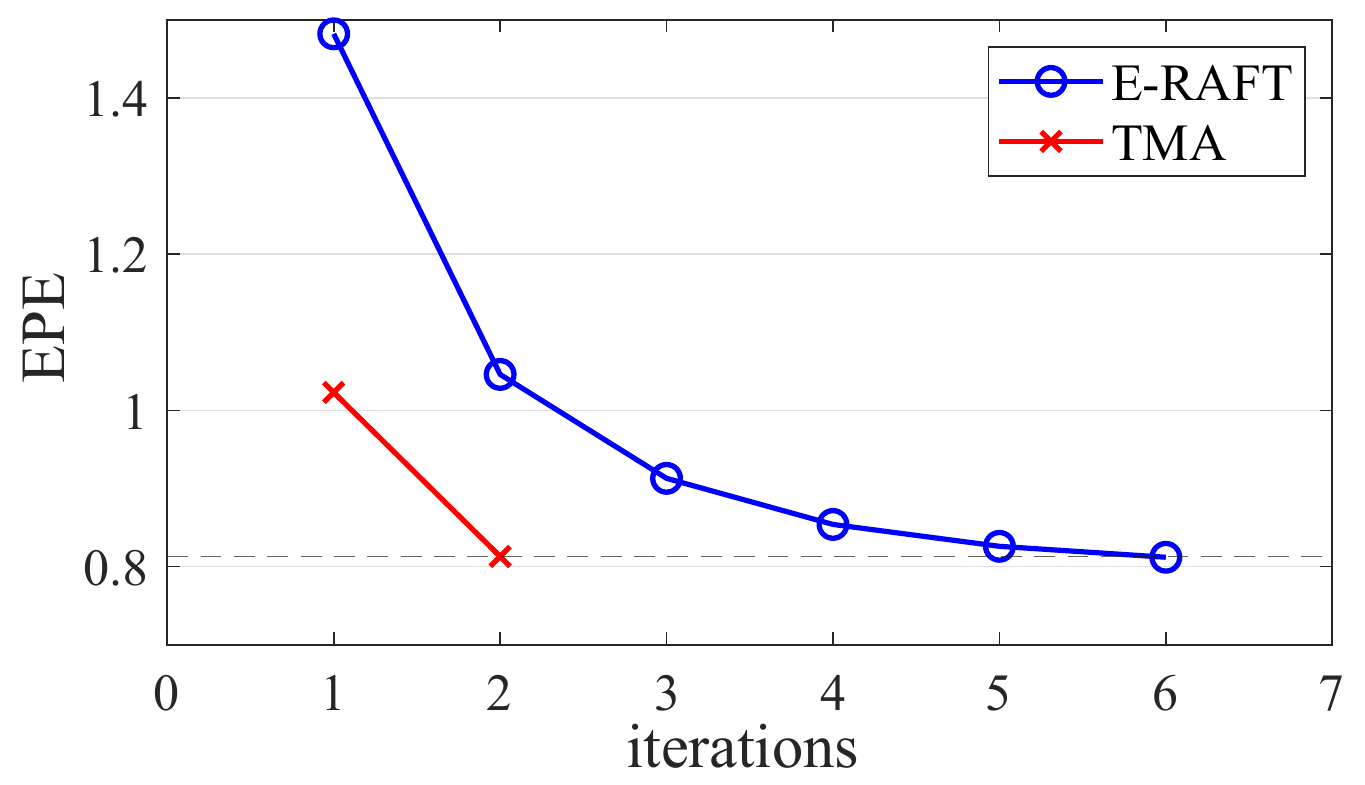}
    \vspace{-0.1in}
    \caption{\textbf{EPE \emph{vs.} number of iterations at inference.} The figure exhibits the prediction results on DSEC-Flow at different stages during inference. Our method achieves an accuracy (EPE = 0.81) comparable to E-RAFT with 2 iterations while E-RAFT needs 6 iterations.}
    \vspace{-0.25in}
    \label{fig:inference}
\end{figure}

\textbf{Efficiency comparison.} The last column of Table~\ref{table:dsec} exhibits the inference time. Our method enjoys faster inference speed compared with E-RAFT, which is mainly attributed to fewer iterative updates. 
The reasonable explanation is that our method utilizes temporal motion information to enrich motion features, thus being able to generate good predictions even at an initial stage. To prove this, we compare the accuracy with E-RAFT at early inference stages in Figure~\ref{fig:inference}. Our method achieves better flow initialization at early stages, supporting our argument.
The ablation experiment on different iterations (Table~\ref{tab:a5}) also demonstrates that our method requires fewer iterations (ours: 6, E-RAFT: 12) to achieve surpassing accuracy. 

\begin{table*}[]
    \centering
    \begin{subtable}[t]{0.495\linewidth}
    \centering
        \vspace{0pt}
        \begin{tabular*}{.95\linewidth}{l@{\extracolsep{\fill}}cccc}
        \toprule
        splits & EPE  &  1PE & 3PE & Time (ms)   \\
        \midrule
        1        &0.79          &11.91    &2.75          &16\\
        3        &0.76          &10.98	&2.45         &18\\
        {\ul 5}  &\textbf{0.74} &\textbf{10.86}	&2.30          &30\\
        7        &\textbf{0.74} &11.13   &\textbf{2.26}  &50\\
        \bottomrule
        \end{tabular*}
        \vspace{-0.05in}
        \caption{\textbf{Event splitting.} Intermediate motion information contributes.}
        \label{tab:a1}
    \end{subtable}
    \begin{subtable}[t]{0.495\linewidth}
    \centering
        \vspace{0pt}
        \begin{tabular*}{.95\linewidth}{l@{\extracolsep{\fill}}ccc}
        \toprule
        style   & EPE  &  1PE & 3PE    \\ 
        \midrule
        w/o     & 0.78         &12.00	&2.66\\
        {\ul linear}  &\textbf{0.74} &\textbf{10.86}	&\textbf{2.30}\\
        same    &0.76	        &11.45	&	2.48\\
        \bottomrule
        \end{tabular*}
        \vspace{-0.05in}
        \caption{\textbf{Lookup style.} Sampling correlation volumes with a set of linear coordinates improves the accuracy.}
        \label{tab:a2}
    \end{subtable}

    \begin{subtable}[t]{0.495\linewidth}
    \centering
        \begin{tabular*}{.95\linewidth}{l@{\extracolsep{\fill}}cccc}
        \toprule
        setup  & EPE  &  1PE & 3PE&Time (ms)\\ 
        \midrule
        {\ul basic}       & \textbf{0.74} &\textbf{10.86}	& 	\textbf{2.30}  &30  \\
        +v\_proj  &  0.75&	10.90&	2.35 &  30 \\
        circular      & 0.76 & 11.25 &2.44 & 30 \\
        \bottomrule
        \end{tabular*}
        \vspace{-0.05in}
        \caption{\textbf{Pattern aggregation module design.}}
        \label{tab:a3}
    \end{subtable}
    \begin{subtable}[t]{0.495\linewidth}
    \centering
        \begin{tabular*}{.95\linewidth}{l@{\extracolsep{\fill}}cccc}
        \toprule
        layers & EPE  &  1PE & 3PE & Time (ms)\\ 
        \midrule
        0        &   0.75    & 11.36      & 2.33 &20        \\
        {\ul 1}        &\textbf{0.74} &\textbf{10.86}	&\textbf{2.30}&30\\
        2  &   0.75    &  10.90	&	2.32&	38\\
        \bottomrule
        \end{tabular*}
        \vspace{-0.05in}
        \caption{\textbf{Layers of pattern aggregation module.} Stacking 1 pattern aggregation module showcases high accuracy.}
        \label{tab:a4}
    \end{subtable}
    
    \begin{subtable}[t]{0.495\linewidth}
    \centering
        \begin{tabular*}{.95\linewidth}{l@{\extracolsep{\fill}}cccc}
        \toprule
        iterations  & EPE  &  1PE   &3PE    & Time (ms)  \\ 
        \midrule
        2           &0.83 &11.04  &2.37  &10         \\
        4           &0.76 &11.45  &2.43  &16      \\
        {\ul 6}     &\textbf{0.74} &10.86	&\textbf{2.30}&30\\
        8           &\textbf{0.74}       &\textbf{10.81}       & 2.32 &54     \\
        \bottomrule
        \end{tabular*}
        \vspace{-0.05in}
        \caption{\textbf{Iterations.} TMA achieves good results with a few iterations.}
        \label{tab:a5}
    \end{subtable}
    \hfill
    \begin{subtable}[t]{0.495\linewidth}
    \centering
        \begin{tabular*}{.95\linewidth}{l@{\extracolsep{\fill}}ccc}
        \toprule
        radius & EPE  &  1PE & 3PE\\ 
        \midrule
        1        & 0.75	&11.14&	2.37   \\
        2        & 0.76 &11.16 &2.39   \\
        {\ul 3}       & \textbf{0.74} &\textbf{10.86}	&	\textbf{2.30}\\
        4       &0.75	&10.93	&\textbf{2.30}          \\
        \bottomrule
        \end{tabular*}
        \vspace{-0.05in}
        \caption{\textbf{Searching radius.} A radius of 3 leads to high accuracy.}
        \label{tab:a6}
    \end{subtable}
    \hfill
\vspace{-0.1in}
\caption{\textbf{TMA ablations.} Models are trained on DSEC-Flow. Settings used in our final model are underlined.}\label{tab:ablation}
\vspace{-0.2in}
\end{table*}

\subsection{MVSEC}
\textbf{Inter-domain evaluation.} Table~\ref{tab:mvsec} reports the results on MVSEC training with outdoor\_day2 sequence. The top of Table~\ref{tab:mvsec} reports the results corresponding to $dt=1$ grayscale frame and the bottom corresponds to $dt=4$ grayscale frames. Our method shows comparable accuracy on outdoor\_day1 for $dt=1$ and highest accuracy on outdoor\_day1 for $dt=4$ among all compared methods. Unfortunately, like two other supervised learning methods, DCEIFlow~\cite{TIP} and E-RAFT~\cite{eraft}, our method fails to achieve superior results on indoor\_flying sequences. One possible reason is that model based methods and unsupervised learning methods usually have better generalization ability, while supervised learning methods are limited if training data has a significantly large gap with test data~\cite{gmflow,TIP}. Since outdoor\_day sequences and indoor\_flying sequences have a great domain gap, it is challenging for supervised learning methods to generalize well on indoor\_flying sequences.

\textbf{Intra-domain evaluation.} To verify the possible cause and better analyze the accuracy of our method under supervised learning setting, we incorporate one indoor\_flying sequence into training set each time and make comparisons with the supervised learning state-of-the-art E-RAFT. Table~\ref{tab:MVSEC_gen} shows the evaluation results corresponding to $dt=4$ grayscale frames (For $dt=1$, refer to appendix). Compared with the results in Table~\ref{tab:mvsec}, E-RAFT and our method both achieve great improvement on indoor\_flying sequences (EPE of indoor\_flying2, ours: 4.32 $\rightarrow$ 1.67, E-RAFT: 5.09 $\rightarrow$ 1.77), confirming the presence of domain gap between outdoor\_day and indoor\_flying data. Compared with the accuracy of E-RAFT with the same training set, on the one hand, our method shows better generalization ability, which generates lower EPE and \% Outlier without seeing indoor\_flying sequences. On the other hand, our method achieves higher accuracy on four sequences by incorporating indoor\_flying sequences into training set. Overall, our method achieves a new state-of-the-art result, demonstrating the effectiveness of our proposed method.

\subsection{Ablations}
We conduct a series of ablation experiments to validate the proposed improvements. All ablation models are trained on DSEC-Flow and evaluated on the public benchmark.

\textbf{Event splitting.} We compare different numbers of event segments in Table~\ref{tab:a1}. We choose different splits in 1, 3, 5 and 7. With the split set to 1, our model degrades into a classic correlation volume based framework and exhibits similar accuracy in comparison with E-RAFT. More splits lead to higher accuracy but lower inference efficiency, so we use 5 splits to achieve accuracy-efficiency trade-off.

\textbf{Lookup style.} Based on the assumption that the optical flow changes linearly over a short period, we propose the linear lookup strategy. To validate it, we try to cancel the lookup operation or just use same sampling coordinates on intermediate correlation volumes. As Table~\ref{tab:a2} shows, sampling the correlation maps with a set of linear coordinates leads to high accuracy.

\textbf{Pattern aggregation module design.} We ablate different components of pattern aggregation module in Table~\ref{tab:a3}. By default, we set the projection matrix of value $W^V$ as an identity matrix because the motion features are encoded by a motion feature encoder beforehand. In addition, we replace the pairing mode of motion features with circular pairing. Circular pairing gives equal treatment to all motion features without highlighting the importance of the last motion feature, which fails to improve accuracy.

\textbf{Layers of pattern aggregation module.} We compare the accuracy with different layers of pattern aggregation modules in Table~\ref{tab:a4}. We note that stacking 1 layer of pattern aggregation module showcases high accuracy. More layers may lead to difficulties in optimization, which in turn causes a decrease in accuracy.

\textbf{Iterations.} Iterative learning plays a role in numerous works~\cite{IRR,raft}.
Results in Table~\ref{tab:a5} show that 6 iterative updates reach the sweet spot of accuracy and efficiency.

\textbf{Searching radius.} We compare different radii for looking up correlation volumes in Table~\ref{tab:a6}. Searching radius of 3 leads to high accuracy. Different from RAFT~\cite{raft} that achieves better predictions with an increased radius, our model does not benefit a lot from a greater radius.

\section{Conclusion}
In this paper, we delve into event-based optical flow estimation problem. Different from existing methods that represent the event stream as static frames to adopt the classic correlation volume based pipeline, we contend that the temporal continuity is a vital element of event-based optical flow.
We propose a novel Temporal Motion Aggregation (TMA) approach to unlock its potential within a learning-based framework. 
By incorporating temporally fine-grained motion information, TMA generates high-quality flow estimates at early stages, which not only enables TMA to obtain accurate final predictions, but also greatly reduces the demand for a number of refinements. 
Extensive experiments on DESC-Flow and MVSEC datasets verify the effectiveness and superiority of TMA.   
Meanwhile, we notice the limitations of supervised learning in generalization ability and expect further improvement to extend our idea into unsupervised learning frameworks.

\noindent\textbf{Acknowledgments}: This work was supported in part by Shanghai Municipal Science and Technology Major Project under Grant 2018SHZDZX01, ZJ Lab, the Shanghai Center for Brain Science and Brain-Inspired Technology, in part by Shanghai Rising Star Program under Grant 21QC1400900, in part by Tongji-Qomolo Autonomous Driving Commercial Vehicle Joint Lab Project, and in part by Xiaomi Young Talents Program. 
{\small
\bibliographystyle{ieee_fullname}
\bibliography{egbib}
}

\end{document}